\documentclass[a4paper,12pt]{article}
\usepackage[dvips]{graphicx}
\usepackage{float}
\usepackage{amsmath} 
\usepackage{amssymb}
\usepackage{amsfonts}
\usepackage{theorem}

\numberwithin{equation}{section}

\newtheorem{rem}{Remark}[section]
\newtheorem{theo}{Theorem}[section]
\newtheorem{lem}{Lemma}[section]

\newtheorem{pro}{Proposition}[section]
\newtheorem{exmp}{Example}[section]
\newtheorem{fact}{Fact}[section]

{
\theoremstyle{break}
\theorembodyfont{\rmfamily}
}

\renewcommand{\thefootnote}{\fnsymbol{footnote}}

\newcommand{\bZ}{\mathbf{Z}}
\newcommand{\bz}{\mathbf{z}}
\newcommand{\bX}{\mathbf{X}}
\newcommand{\bx}{\mathbf{x}}

\newcommand{\prob}{\mathbb{P}}
\newcommand{\esp}{\mathbb{E}}
\newcommand{\ind}{\mathbf{1}}
\newcommand{\card}[1]{|{#1}|}
\renewcommand{\L}{\mathcal L}
\newcommand{\R}{\mathbb R}
\newcommand{\de}{\mathrm d}

\newif\ifnotes
\notestrue
\newcounter{mnotecount}[section]

\begin{document}

\begin{center}

{\sc \Large Statistical Analysis of $k$-Nearest Neighbor Collaborative Recommendation\\
\vspace{0.7cm}}

G\'erard BIAU $^{\mbox{\footnotesize a,}}\footnote{Corresponding
author.}$, Beno\^{i}t CADRE $^{\mbox{\footnotesize b}}$ and
Laurent ROUVI\`ERE $^{\mbox{\footnotesize c}}$

\vspace{0.5cm}

$^{\mbox{\footnotesize a}}$ LSTA \& LPMA\\Universit\'e Pierre et Marie Curie -- Paris VI\\
Bo\^{\i}te 158,   175 rue du Chevaleret\\
75013 Paris, France\\
\smallskip
\textsf{gerard.biau@upmc.fr}\\
\bigskip
$^{\mbox{\footnotesize b}}$ 
IRMAR, ENS Cachan Bretagne, CNRS, UEB\\
Campus de Ker Lann\\
Avenue Robert Schuman\\
35170 Bruz, France\\
\smallskip
\textsf{Benoit.Cadre@bretagne.ens-cachan.fr}\\
\bigskip
$^{\mbox{\footnotesize c}}$ CREST-ENSAI, IRMAR, UEB\\
Campus de Ker Lann\\
Rue Blaise Pascal - BP 37203\\
35172 Bruz Cedex, France\\
\smallskip
\textsf{laurent.rouviere@ensai.fr}\\

\vspace{0.5cm}

\end{center}

\begin{abstract} 
\noindent {\rm Collaborative recommendation is an information-filtering technique that attempts to present information items that are likely of interest to an Internet user. Traditionally, collaborative systems deal with situations with two types of variables, users and items. In its most common form, the problem is framed as trying to estimate ratings for items that have not yet been consumed by a user. Despite wide-ranging literature, little is known about the statistical properties of recommendation systems. In fact, no clear probabilistic model even exists which would allow us to precisely describe the mathematical forces driving collaborative filtering. To provide an initial contribution to this, we propose to set out a general sequential stochastic model for collaborative recommendation. We offer an in-depth analysis of the so-called cosine-type nearest neighbor collaborative method, which is one of the most widely used algorithms in collaborative filtering, and analyze its asymptotic performance as the number of users grows.  We establish consistency of the procedure under mild assumptions on the model. Rates of convergence and examples are also provided. 
\medskip

\medskip
 
}

\end{abstract}

\renewcommand{\thefootnote}{\arabic{footnote}}

\setcounter{footnote}{0}
\section{Introduction}
Collaborative recommendation is a Web information-filtering technique that typically gathers information about your personal interests and compares your profile to other users with similar tastes. The goal of this system is to give personalized recommendations, whether this be movies you might enjoy, books you should read or the next restaurant you should go to.
\bigskip

There has been much work done in this area over the past decade since the appearance of the first papers on the subject in the mid-90's (Resnick et al. \cite{RISBR94}, Hill et al. \cite{HSRF95}, Shardanand and Maes \cite{SM95}). Stimulated by an abundance of practical applications, most of the research activity to date has focused on elaborating various heuristics and practical methods (Breese et al. \cite{BHK1998}, Heckerman et al. \cite{HCMRK2000}, Salakhutdinov et al. \cite{SMH2007}) so as to provide personalized recommendations and help Web users deal with information overload. Examples of such applications include recommending books, people, restaurants, movies, CDs and news. Websites such as amazon.com, match.com, movielens.org and allmusic.com already have recommendation systems in operation. We refer the reader to the surveys by Adomavicius and Tuzhilin \cite{AT2005} and Adomavicius et al. \cite{ASST2005} for a broader picture of the field, an overview of results and many related references.
\bigskip

Traditionally, collaborative systems deal with situations with two types of variables, {\it users} and {\it items}. In its most common form, the problem is framed as trying to estimate {\it ratings} for items that have {\it not} yet been consumed by a user. The recommendation process typically starts by asking users a series of questions about items they liked or did not like. For example, in a movie recommendation system, users initially rate some subset of films they have already seen. Personal ratings are then collected in a matrix, where each row represents a user, each column an item, and entries in the matrix represent a given user's rating of a given item. An example is presented in Table \ref{tableau1}, where ratings are specified on a scale from 1 to 10, and  ``NA''  means that the user has not rated the corresponding film. 
\medskip

\begin{table}[!h]
\begin{center}
\begin{tabular}{|c|c|c|c|c|c|c|}
\hline
\hline
& {\small Armageddon} 
& {\small Platoon} & {\small Rambo} & {\small Rio Bravo} & {\small Star wars} & {\small Titanic} \\
\hline
{\small Jim} & NA & 6 & 7 & 8 & 9 & NA\\
{\small James} & 3 & NA & 10 & NA & 5 & 7\\
{\small Steve} & 7 & NA & 1 & NA & 6 & NA\\
{\small Mary} & NA & 7 & 1 & NA & 5 & 6\\
{\small John} & NA & 7 & NA & NA & 3 & 1\\
{\small Lucy} & 3 & 10 & 2 & 7 & NA & 4\\
{\small Stan} & NA & 7 & NA & NA & 1 & NA\\
{\small Johanna} & 4 & 5 & NA & 8 & 3 & 9\\
\hline
\hline
{\small Bob} & NA & 3 & 3 & 4 & 5 & $\mathbf{?}$\\
\hline
\hline
\end{tabular}
\caption{A (subset of a) ratings matrix for a movie recommendation system. Ratings are specified on a scale from 1 to 10, and  ``NA''  means that the user has not rated the corresponding film.}
\label{tableau1}
\end{center}
\end{table}
Based on this prior information, the recommendation engine must be able to automatically furnish ratings of as-yet unrated items and then suggest appropriate recommendations based on these predictions.  To do this, a number of practical methods have been proposed, including machine learning-oriented techniques (e.g., Abernethy et al. \cite{ABEV2008}), statistical approaches (e.g., Sarwar et al. \cite{SKKR2001}) and numerous other ad hoc rules (Adomavicius and Tuzhilin \cite{ASST2005}). The collaborative filtering issue may be viewed as a special instance of the problem of inferring the many missing entries of a data matrix. This field, which has very recently emerged, is known as the matrix completion problem, and comes up in many areas of science and engineering, including collaborative filtering, machine learning, control, remote sensing and computer vision. We will not pursue this promising approach, and refer the reader to Cand\`es and Recht  \cite{candes2} and Cand\`es and Plan \cite{candes1} who survey the literature on matrix completion. These authors show in particular that under suitable conditions, one can recover an unknown low rank matrix from a nearly minimal set of entries by solving a simple convex optimization problem.
\bigskip

In most of the approaches, the crux is to identify users whose tastes/ratings are ``similar'' to the user we would like to advise. The similarity measure assessing proximity between users may vary depending on the type of application, but is typically based on a correlation or cosine-type approach (Sarwar et al. \cite{SKKR2001}).
\bigskip

Despite wide-ranging literature, very little is known about the statistical properties of recommendation systems. In fact, no clear probabilistic model even exists allowing us to precisely describe the mathematical forces driving collaborative filtering. To provide an initial contribution to this, we propose in the present paper to set out a general stochastic model for collaborative recommendation and analyze its asymptotic performance as the number of users grows. 
\bigskip

The document is organized as follows. In section 2, we provide a sequential stochastic model for collaborative recommendation and describe the statistical problem. In the model we analyze, unrated items are estimated by averaging ratings of users who are ``similar'' to the user we would like to advise. The similarity is assessed by a cosine-type measure, and unrated items are estimated using a $k_n$-nearest neighbor-type regression estimate, which is indeed one of the most widely used procedures in collaborative filtering. It turns out that the choice of the cosine proximity as a similarity measure imposes constraints on the model, which are discussed in section 3.  Under mild assumptions, consistency of the estimation procedure is established in section 4, whereas rates of convergence are discussed in section 5. Illustrative examples are given throughout the document, and proofs of some technical results are postponed to section 6.

\section{A model for collaborative recommendation}
\subsection{Ratings matrix and new users}
Suppose that there are $d+1$ ($d \geq 1$) possible items, $n$ users in the ratings matrix (i.e., the database) and that users' ratings take values in the set $(\{0\} \cup [1,s])^{d+1}$. Here, $s$ is a real number greater than 1 corresponding to the maximal rating and, by convention, the symbol 0 means that the user has not rated the item (same as ``NA''). Thus, the ratings matrix has $n$ rows, $d+1$ columns and entries from $\{0\} \cup [1,s]$. For example, $n=8$, $d=5$ and $s=10$ in Table \ref{tableau1}, which will be our toy example throughout this section. Then, a new user Bob reveals some of his preferences for the first time, rating some of the first $d$ items but {\it not} the $(d+1)$th (the movie Titanic in Table \ref{tableau1}). We want to design a strategy to predict Bob's rating of Titanic using: $(i)$ Bob's ratings of
some (or all) of the other $d$ movies and $(ii)$ the ratings matrix. This is illustrated in Table \ref{tableau1}, where Bob has rated 4 out of the 5 movies.
\bigskip

The first step in our approach is to model the preferences of new user Bob by a random vector $(\bX,Y)$ of size $d+1$ taking values in the set $[1,s]^d \times [1,s]$. Within this framework, the random variable $\bX=(X_1, \hdots, X_d)$ represents Bob's preferences pertaining to the first $d$ movies, whereas $Y$, the (unobserved) variable of interest, refers to the movie Titanic. In fact, as Bob does not necessarily reveals all his preferences at once, we do not observe the variable $\bX$, but instead some ``masked'' version of it denoted hereafter by $\bX^{\star}$. The random variable $\bX^{\star}=(X^{\star}_1, \hdots, X^{\star}_d)$ is naturally defined by
\begin{equation*}
\label{joletaxi}
X_j^{\star}=
\left \{ \begin{array}{cl}
X_j& \mbox{if } j \in M\\
0 & \mbox{otherwise},
\end{array}
\right .
\end{equation*}
where  $M$ stands for some non-empty random subset of $\{1,\hdots, d\}$ indexing the  movies which have been rated by Bob. Observe that the random variable $\bX^{\star}$ takes values in $(\{0\}\cup [1,s])^{d}$ and that $\|\bX^{\star}\| \geq 1$, where $\|.\|$ denotes the usual Euclidean norm on $\mathbb R^d$. In the example of Table \ref{tableau1}, $M=\{2,3,4,5\}$ and (the realization of) $\bX^{\star}$ is $(0,3,3,4,5)$.
\bigskip

We follow the same approach to model preferences of users already in the database (Jim,  James, Steve, Mary, etc. in Table \ref{tableau1}), who will therefore be represented by a sequence of independent $[1,s]^d \times [1,s]$-valued random pairs $(\bX_1,Y_1), \hdots, (\bX_n,Y_n)$ from the distribution $(\bX,Y)$. A first idea for dealing with potential non-responses of a user $i$ in the ratings matrix ($i=1, \hdots, n$) is to consider in place of $\bX_i=(X_{i1}, \hdots, X_{id})$ its masked version $\widetilde{\bX}_i=(\widetilde{X}_{i1}, \hdots, \widetilde{X}_{id})$ defined by
\begin{equation}
\label{crise}
\widetilde{X}_{ij}=
\left \{ \begin{array}{cl}
X_{ij}& \mbox{if } j \in M_i \cap M\\
0 & \mbox{otherwise}, 
\end{array}
\right .
\end{equation}
where each $M_i$ is the random subset of $\{1,\hdots,d\}$ indexing the movies which have been rated by user $i$. In other words, we only keep in $\bX_i$ items corated by both user $i$ {\it and} the new user --- items which have not been rated by $\bX$ and $\bX_i$ are declared non-informative and simply thrown away.
\bigskip

However, this model, which is static in nature, does not allow to take into account the fact that, as time goes by, each user in the database may reveal more and more preferences. This will for instance typically be the case in the movie recommendation system of Table \ref{tableau1}, where regular customers will update their ratings each time they have seen a new movie. Consequently, model (\ref{crise}) is not fully satisfying and must therefore be slightly modified to better capture the sequential evolution of ratings.

\subsection{A sequential model}
A possible dynamical approach for collaborative recommendation is based on the following protocol: users enter the database one after the other and update their list of ratings sequentially in time. More precisely, we suppose that at each time $i=1, 2, \hdots$, a new user enters the process and reveals his preferences for the first time, while the $i-1$ previous users are allowed to rate new items. Thus, at time 1, there is only one user in the database (Jim in Table \ref{tableau1}), and the (non-empty) subset of items he decides to rate is modeled by a random variable $M_1^1$ taking values in $\mathcal P^{\star}(\{1, \hdots, d\})$, the set of non-empty subsets of $\{1,\hdots,d\}$. At time 2, a new user (James) enters the game and reveals his preferences according to a $\mathcal P^{\star}(\{1, \hdots, d\})$-valued random variable $M_2^1$, with the same distribution as $M_1^1$. At the same time, Jim (user 1) may update his list of preferences, modeled by a random variable $M_1^2$ satisfying $M_1^1\subset M_1^2$. The latter requirement just means that the user is allowed to rate new items but not to remove his past ratings.  At time 3, a new user (Steve) rates items according to a random variable $M_3^1$ distributed as $M_1^1$, while user 2 updates his preferences according to $M_2^2$ (distributed as $M_1^2$) and user 1 updates his own according to $M_1^3$, and so on.  This sequential mechanism is summarized in Table \ref{tableau2}.
\medskip

\begin{table}[!h]
\begin{center}
\begin{tabular}{|c|c|c|c|c|c|c|}
\hline
\hline
 & Time 1 & Time 2 & $\hdots$ & Time $i$ & $\hdots$ & Time $n$\\
 \hline
User 1 & $M_1^1$ & $M_1^2$ & $\hdots$ & $M_1^i $ & $\hdots$ & $M_1^n$ \\
User 2 &  & $M_2^1$ & $\hdots$ & $M_2^{i-1} $ & $\hdots$ & $M_2^{n-1}$ \\
\vdots & &&$\ddots$&$\vdots$& $\vdots$& $\vdots$\\
User $i$ & && & $M_i^1$ & $\hdots$ & $M_i^{n+1-i}$\\
\vdots & &&&& $\ddots$ &$\vdots$\\
User $n$ & &&&&&$M_n^1$\\
\hline
\hline
\end{tabular}
\caption{A sequential model for preference updating.}
\label{tableau2}
\end{center}
\end{table}

By repeating this procedure, we end up at time $n$ with an upper triangular array $(M_i^j)_{1\leq i \leq n, 1\leq j \leq n+1-i}$ of random variables. A row in this array consists of a collection $M_{i}^j$ of random variables for a given value of $i$, taking values in $\mathcal P^{\star}(\{1, \hdots, d\})$ and satisfying the constraint $M_{i}^j \subset M_{i}^{j+1}$.  For a fixed $i$, the sequence $M_{i}^1 \subset M_{i}^{2}\subset \hdots$ describes the (random) way user $i$ sequentially reveals his preferences over time. Observe that the later inclusions are not necessarily strict, so that a single user is not forced to rate one more item at every single step.
\bigskip

Throughout the paper, we will assume that, for each $i$, the distribution of the sequence of random variables $(M_i^n)_{n\geq 1}$ is independent of $i$, and is therefore distributed as a generic random sequence denoted $(M^n)_{n\geq 1}$,  satisfying $M^1\neq \emptyset$ and $M^n \subset M^{n+1}$ for all $n \geq 1$. For the sake of coherence, we assume that $M^1$ and $M$ (see (\ref{joletaxi})) have the same distribution, i.e.,  the new abstract user $\bX^{\star}$ may be regarded as a user entering the database for the first time. We will also suppose that there exists a positive random integer $n_0$ such that $M^{n_0}=\{1, \hdots, d\}$ and, consequently, $M^n=\{1, \hdots, d\}$ for all $n \geq n_0$. This requirement means that each user rates all $d$ items after a (random) period of time. Last, we will assume that the pairs $(\bX_i,Y_i)$, $i=1, \hdots, n$, the sequences $(M_1^n)_{n \geq 1}$, $(M_2^n)_{n \geq 1}, \hdots$ and the random variable $M$ are mutually independent. We note that this implies that the users' ratings are independent.
\bigskip

With this sequential point of view, improving on (\ref{crise}), we let the masked version $\bX_i^{(n)}=(X_{i1}^{(n)}, \hdots, X_{id}^{(n)})$ of $\bX_i$ be defined as
\begin{equation*}
\label{crise2}
X_{ij}^{(n)}=
\left \{ \begin{array}{cl}
X_{ij}& \mbox{if } j \in M_{i}^{n+1-i} \cap M\\
0 & \mbox{otherwise}.
\end{array}
\right .
\end{equation*}
Again, it is worth pointing out that, in the definition of $\bX_i^{(n)}$, items which have not been corated by both $\bX$ {\it and} $\bX_i$ are deleted. This implies in particular that $\bX_i^{(n)}$ may be equal to $\bf{0}$, the $d$-dimensional null vector (whereas $\|\bX^{\star}\|\geq 1$ by construction).
\bigskip

Finally, in order to deal with possible non-answers of database users regarding the variable of interest  (Titanic in our movie example), we introduce $(\mathcal R_n)_{n \geq 1}$, a sequence of random variables taking values in $\mathcal P^{\star}(\{1, \hdots, n\})$, such that $\mathcal R_n$ is independent of $M$ and the sequences $(M_i^{n})_{n \geq 1}$, and satisfying $\mathcal R_n \subset \mathcal R_{n+1}$ for all $n \geq 1$. In this formalism, $\mathcal R_n$ represents the subset, which is assumed to be non-empty, of users who have already provided information about Titanic at time $n$. For example, in Table \ref{tableau1}, only James, Mary, John, Lucy and Johanna have rated Titanic and therefore (the realization of) $\mathcal R_n$ is $\{2,4,5,6,8\}$.
\subsection{The statistical problem}
To summarize the model so far, we have at hand at time $n$ a sample of random pairs $(\bX_1^{(n)}, Y_1), \hdots, (\bX_n^{(n)},Y_n)$ and our mission is to predict the score $Y$ of a new user represented by $\bX^\star$. The variables $\bX_1^{(n)}, \hdots, \bX_n^{(n)}$ model  the database users' revealed preferences with respect to the first $d$ items. They take values in $(\{0\}\cup [1,s])^d$, where a 0 at coordinate $j$ of $\bX_i^{(n)}$ means that the $j$th product has not been corated by both user $i$ and the new user. The variable $\bX^\star$ takes values in $(\{0\}\cup [1,s])^d$ and satisfies $\| \bX^{\star}\|\geq 1$. The random variables $Y_1, \hdots, Y_n$ model users' ratings of the product of interest. They take values in $[1,s]$ and, at time $n$, we only see a non-empty (random) subset of $\{Y_1, \hdots, Y_n\}$, indexed by $\mathcal R_n$.
\bigskip

The statistical problem with which we are faced is to estimate the regression function $\eta(\bx^{\star})=\mathbb E[Y|\bX^{\star}=\bx^{\star}]$. For this goal, we may use the database observations $(\bX_1^{(n)}, Y_1), \hdots, (\bX_n^{(n)},Y_n)$ in order to construct an estimate $\eta_n(\bx^{\star})$ of $\eta(\bx^{\star})$. The approach we explore in this paper is a cosine-based $k_n$-nearest neighbor regression method, one of the most widely used algorithms in collaborative filtering (e.g., Sarwar et al. \cite{SKKR2001}).
\bigskip

Given $\bx^{\star} \in (\{0\}\cup [1,s])^d-{\bf 0}$ and the sample $(\bX_1^{(n)}, Y_1), \hdots, (\bX_n^{(n)},Y_n)$,  the idea of the cosine-type $k_n$-nearest neighbor (NN) regression method is to estimate $\eta(\bx^{\star})$ by a local averaging over those $Y_i$ for which: ($i$) $\bX_i^{(n)}$ is ``close'' to $\bx^\star$ and ($ii$) $i \in \mathcal R_n$, that is, we effectively ``see'' the rating $Y_i$. For this, we scan through the $k_n$ neighbors of $\bx^{\star}$ among the database users $\bX_i^{(n)}$ for which $i \in \mathcal R_n$ and estimate $\eta(\bx^{\star})$ by averaging the $k_n$ corresponding $Y_i$. The closeness between users is assessed by a cosine-type similarity, defined for $\bx=(x_1, \hdots, x_d)$ and $\bx'=(x'_1, \hdots, x'_d)$ in $(\{0\}\cup [1,s])^d$ by
$$\bar S(\bx, \bx')=\frac{\sum_{j\in\mathcal J}x_jx_j'}{\sqrt{\sum_{j\in\mathcal J}x_j^2}\sqrt{\sum_{j\in\mathcal J}x_j'^2}},$$
where $\mathcal J=\{j\in\{1,\hdots,d\}:x_j\neq 0\textrm{ and }x_j'\neq 0\}$ and, by convention, $\bar S(\bx, \bx')=0$ if $\mathcal J=\emptyset$. To understand the rationale behind this proximity measure, just note that if $\mathcal J=\{1,\hdots,d\}$ then $\bar S(\bx, \bx')$ coincides with $\cos(\bx, \bx')$, i.e.,  two users are ``close'' with respect to $\bar S$ if their ratings are more or less proportional. However, the similarity $\bar S$, which will be used to measure the closeness between $\bX^\star$ (the new user) and $\bX_i^{(n)}$ (a database user) ignores possible non-answers in $\bX^{\star}$ or $\bX_i^{(n)}$, and is therefore more adapted to the recommendation setting. For example, in Table \ref{tableau1}, 
$$\bar S(\mbox{Bob}, \mbox{Jim})=\bar S ((0,3,3,4,5),(0,6,7,8,9))=\bar S((3,3,4,5),(6,7,8,9))\approx 0.99,$$
whereas 
$$\bar S(\mbox{Bob},\mbox{Lucy})=\bar S((0,3,3,4,5),(3,10,2,7,0))=\bar S((3,3,4),(10,2,7))\approx 0.89.$$
Next, fix $\bx^{\star} \in (\{0\}\cup [1,s])^d-{\bf 0}$ and suppose to simplify that $M\subset M_{i}^{n+1-i}$ for each $i\in \mathcal R_n$. In this case, it is easy to see that $\bX_i^{(n)}= \bX_i^{\star}=(X_{i1}^{\star}, \hdots,X_{id}^{\star})$, where
\begin{equation*}
X_{ij}^{\star}=
\left \{ \begin{array}{cl}
X_{ij}& \mbox{if } j \in M\\
0 & \mbox{otherwise}.
\end{array}
\right .
\end{equation*}
Besides, $Y_i \geq 1$, 
\begin{equation}
\label{isup}
\bar S(\bx^{\star},\bX_i^{\star})=\cos(\bx^{\star},\bX_i^{\star})>0,
\end{equation}
 and an elementary calculation shows that the positive real number $y$ which maximizes the similarity between $(\bx^{\star},y)$ and $(\bX_i^{\star},Y_i)$, that is  
$$\bar S\left((\bx^{\star},y),(\bX_i^{\star},Y_i)\right)=\frac{\sum_{j\in M}x_j^\star X_{ij}^\star+yY_i}{\sqrt{\sum_{j\in M}{x_j^\star}^2+y^2}\sqrt{\sum_{j\in M}{X_{ij}^\star}^2+Y_i^2}},$$
is given by
$$y=\frac{\|\bx^{\star}\|}{\|\bX_i^{\star}\| \cos(\bx^{\star},\bX_i^{\star})}\,Y_i.$$
This suggests the following regression estimate $\eta_n(\bx^{\star})$ of $\eta(\bx^{\star})$:
\begin{equation}
\label{benoitcadreboittrop}
\eta_n(\bx^{\star})=\| \bx ^{\star} \| \sum_{i \in \mathcal R_n} W_{ni}(\bx^{\star}) \frac{Y_i}{\|\bX_i^{\star} \|},
\end{equation}
where the integer $k_n$ satisfies $1\leq k_n \leq n$ and
\begin{displaymath}
W_{ni}(\bx^{\star})= \left \{
\begin{array}{ll}
1/k_n & \mbox{if } \bX_i^{\star} \mbox{ is among the $k_n$-MS of } \bx^{\star} \mbox{ in } \{\bX_i^{\star}, i \in \mathcal R_n\}\\
 0 & \mbox{otherwise.}
 \end{array}
 \right .
\end{displaymath}
In the above definition, the acronym ``MS'' (for Most Similar) means that we are searching for the $k_n$ ``closest'' points of $\bx^{\star}$ within the set $\{\bX_i^{\star}, i \in \mathcal R_n\}$ using the similarity $\bar S$ --- or, equivalently here, using the cosine proximity (by identity (\ref{isup})). Note that the cosine term has been removed since it has asymptotically no influence on the estimate, as can be seen by a slight adaptation of the arguments of the proof of Lemma 6.1, Chapter 6, in Gy\"orfi et al. \cite{gykokrha}.
The estimate $\eta_n(\bx^{\star})$ is called the {\it cosine-type $k_n$-NN regression estimate} in the collaborative filtering literature. Now, recalling that definition (\ref{benoitcadreboittrop}) makes sense only when $M\subset M_{i}^{n+1-i}$ for each $i \in \mathcal R_n$ (that is, $\bX_i^{(n)}= \bX_i^{\star}$), the next step is to extend the definition of $\eta_n(\bx^{\star})$ to the general case. In view of (\ref{benoitcadreboittrop}), the most natural approach is to simply put
\begin{equation}
\label{benoitcadreboitvraimenttrop}
\eta_n(\bx^{\star})=\| \bx ^{\star} \| \sum_{i \in \mathcal R_n} W_{ni}(\bx^{\star}) \frac{Y_i}{\|\bX_i^{(n)} \|},
\end{equation}
where 
\begin{displaymath}
W_{ni}(\bx^{\star})= \left \{
\begin{array}{ll}
1/k_n & \mbox{if } \bX_i^{(n)} \mbox{ is among the $k_n$-MS of } \bx^{\star} \mbox{ in } \{\bX_i^{(n)}, i \in \mathcal R_n\}\\
 0 & \mbox{otherwise.}
 \end{array}
 \right .
\end{displaymath}
The acronym ``MS'' in the weight $W_{ni}(\bx^{\star})$ means that the $k_n$ closest database points of $\bx^{\star}$ are computed according to the similarity
$$S\left(\bx^\star,\bX_i^{(n)}\right)=p_i^{(n)}\bar S\left (\bx^\star,\bX_i^{(n)}\right),\quad \mbox{with } p_i^{(n)}=\frac{\card{M_i^{n+1-i}\cap M}}{\card{M}}$$
(here and throughout, notation $\card{A}$ means the cardinality of the finite set $A$). The factor $p_i^{(n)}$ in front of $\bar S$ is a penalty term which, roughly, avoids to over-promote the last users entering the database. Indeed, the effective number of items rated by these users will be eventually low and, consequently, their $\bar S$-proximity to $\bx^{\star}$ will tend to remain high. On the other hand, for fixed $i$ and $n$ large enough, we know that $M\subset M_{i}^{n+1-i}$ and $\bX_i^{(n)}=\bX_i^{\star}$. This implies $p_i^{(n)}=1$, $S(\bx^\star,\bX_i^{(n)})=\bar S(\bx^\star,\bX_i^{\star})=\cos(\bx^\star,\bX_i^{\star})$ and shows that definition (\ref{benoitcadreboitvraimenttrop}) generalizes definition (\ref{benoitcadreboittrop}). Therefore, we take the liberty to still call the estimate (\ref{benoitcadreboitvraimenttrop}) the cosine-type $k_n$-NN regression estimate.
\begin{rem} A smoothed version of the similarity $S$ could also be considered, typically
$$S\left(\bx^\star,\bX_i^{(n)}\right)=\psi\left(p_i^{(n)}\right)\bar S\left(\bx^\star,\bX_i^{(n)}\right),$$
where $\psi\, :\, [0,1]\to [0,1]$ is a nondecreasing map satisfying $\psi(1/2)<1$ (assuming $\card{M}\geq 2$). For example, the choice $\psi(p)=\sqrt {p}$ tends to promote users with a low number of rated items, provided the items corated by the new user are quite similar. In the present paper, we shall only consider the case $\psi(p)=p$, but the whole analysis carries over without difficulties for general functions $\psi$.
\end{rem}

\begin{rem}
Another popular approach to measure the closeness between users is the Pearson correlation coefficient.
The extension of our results to Pearson-type similarities is not straightforward and more work is needed to address this challenging question. We refer the reader to Choi et al. \cite{CKJ06} and Montaner et al. \cite{MLR03} for a comparative study and comments on the choice of the similarity.
\end{rem}

Finally, for definiteness of the estimate $\eta_n(\bx^{\star})$, some final remarks are in order:
\begin{enumerate}
\item[$(i)$]If $\bX_i^{(n)}$ and $\bX_j^{(n)}$ are equidistant from $\bx^{\star}$, i.e., $S(\bx^{\star},\bX_i^{(n)})=S(\bx^{\star},\bX_j^{(n)})$, then we have a tie and, for example, $\bX_i^{(n)}$ may be declared ``closer'' to $\bx^\star$ if $i<j$, that is, tie-breaking is done by indices.
\item[$(ii)$] If $|\mathcal R_n|<k_n$, then the weights $W_{ni}(\bx^{\star})$ are not defined. In this case, we conveniently set $W_{ni}(\bx^{\star})=0$, i.e., $\eta_n(\bx^{\star})=0$. 
\item[$(iii)$] If $\bX_i^{(n)}=\mathbf 0$, then we take $W_{ni}(\bx^{\star})=0$ and we adopt  the convention $0\times \infty=0$ for the computation of $\eta_n(\bx^{\star})$.
\item[$(iv)$] With the above conventions, the identity $\sum_{i \in \mathcal R_n}W_{ni}(\bx^{\star})\leq 1$ holds in each case. 
\end{enumerate}
\section{The regression function}
Our objective in section 4 will be to establish consistency of the estimate $\eta_n(\bx^{\star})$ defined in (\ref{benoitcadreboitvraimenttrop}) towards the regression function $\eta({\bf x}^{\star})$. To reach this goal, we first need to analyze the properties of $\eta({\bf x}^{\star})$. Surprisingly, the special form of $\eta_n(\bx^{\star})$ constrains the shape of $\eta(\bx^{\star})$. This is stated in Theorem \ref{forme} below.
\begin{theo}
\label{forme}
Suppose that $\eta_n({\bf X}^\star) \to \eta({\bf X}^{\star})$ in probability as $n \to \infty$. Then 
$$\eta({\bf X}^\star)=\|{\bf X}^\star\| \, \mathbb E\left[ \frac{Y}{\|{\bf X}^\star\|} \Big| \frac{{\bf X}^\star}{\|{\bf X}^\star\|} \right]\quad  \mbox{a.s.}$$
\end{theo}

\noindent {\bf Proof of Theorem \ref{forme}.}\quad Recall that
$$\eta_n({\bf X}^{\star})=\|{\bf X}^{\star}\|\sum_{i\in \mathcal R_n} W_{ni}({\bf X}^\star) \frac{Y_i}{\|{\bf X}_i^{(n)}\|},$$
and let
$$\varphi_n({\bf X}^\star)=\sum_{i\in \mathcal R_n} W_{ni}({\bf X}^\star) \frac{Y_i}{\|{\bf X}_i^{(n)}\|}.$$
Since $(\eta_n({\bf X}^\star))_n$ is a Cauchy sequence in probability and $\|{\bf X}^\star\|\geq 1$, $(\varphi_n({\bf X}^\star))_n$ is also a Cauchy sequence. Thus, there exists a measurable function $\varphi$ on $\mathbb{R}^d$ such that  $\varphi_n({\bf X}^\star)\to \varphi({\bf X}^\star)$ in probability. Using the fact that $0 \leq \varphi_n({\bf X}^\star) \leq s$ for all $n\geq 1$, we conclude that $0 \leq \varphi({\bf X}^\star)\leq s$ a.s. as well.
\bigskip

Let us extract a sequence $(n_k)_k$ satisfying $\varphi_{n_k}({\bf X}^\star)\to\varphi({\bf X}^\star)$ a.s. Observing that, for ${\bf x}^\star\neq \mathbf 0$, 
$$\varphi_{n_k}({\bf x}^\star)=\varphi_{n_k}\left(\frac{{\bf x}^\star}{\|{\bf x}^\star\|}\right),$$
we may write $\varphi({\bf X}^\star)=\varphi({\bf X}^\star/\|{\bf X}^\star\|)$ a.s. Consequently, the limit in probability of $(\eta_n({\bf X}^\star))_n$ is 
$$\|{\bf X}^\star\| \, \varphi\left(\frac{{\bf X}^\star}{\|{\bf X}^\star\|}\right).$$  
Therefore, by the uniqueness of the limit, $\eta(\bf X^{\star})=\|{\bf X}^\star\|\,\varphi ({\bf X}^\star/\|{\bf X}^\star\|)$ a.s. Moreover, 
\begin{align*}
 \varphi\left(\frac{{\bf X}^\star}{\|{\bf X}^\star\|}\right) & = \mathbb E\left[\varphi\left(\frac{{\bf X}^\star}{\|{\bf X}^\star\|}\right)\Big|\frac{{\bf X}^\star}{\|{\bf X}^\star\|}\right]\\
&  =  \mathbb E \left[\frac{\eta({\bf X}^\star)}{\|{\bf X}^\star\|}\Big|\frac{{\bf X}^\star}{\|{\bf X}^\star\|} \right]\\
& = \mathbb E \left [ \mathbb E \left [\frac{Y}{\|{\bf X}^\star\|} \Big| {\bf X}^\star\right ]\Big | \frac{{\bf X}^\star}{\|{\bf X}^\star\|} \right]\\
& =  \mathbb E \left [ \frac{Y}{\|{\bf X}^\star\|} \Big| \frac{{\bf X}^\star}{\|{\bf X}^\star\|} \right],
\end{align*}
since $\sigma({\bf X}^\star/\|{\bf X}^\star\|)\subset \sigma({\bf X}^\star)$. This concludes the proof of the theorem.
\begin{flushright}
$\square$
\end{flushright}
An important consequence of Theorem \ref{forme} is that if we intend to prove any consistency result regarding the estimate $\eta_n(\mathbf x^{\star})$, then we have to assume that the regression function $\eta({\mathbf x}^{\star})$ has the special form 
\begin{equation*}
\label{F}
\eta({\bf x}^\star)=\|{\bf x}^\star\| \varphi(\bx^\star), \quad\textrm{where}\quad\varphi(\bx^\star)= \mathbb E\left[ \frac{Y}{\|{\bf X}^\star\|} \Big| \frac{{\bf X}^\star}{\|{\bf X}^\star\|}=\frac{{\bf x}^\star}{\|{\bf x}^\star\|} \right] \quad (\mathbf F).
\end{equation*}
This will be our fundamental requirement throughout the paper, and it will be denoted by $(\mathbf F)$. In particular, if $\tilde\bx^\star=\lambda\bx^\star$ with $\lambda >0$, then $\eta(\tilde\bx^\star)=\lambda\eta(\bx^\star)$. That is, if two ratings $\bx^\star$ and $\tilde\bx^\star$ are proportional, then so must be the values of the regression function at $\bx^\star$ and $\tilde\bx^\star$, respectively. 
\section{Consistency}
In this section, we establish the $L_1$ consistency of the regression estimate $\eta_n(\bx^{\star})$ towards the regression function $\eta(\bx^{\star})$. Using $L_1$ consistency is essentially a matter of taste, and all the subsequent results may be easily adapted to $L_p$ norms without too much effort. In the proofs, we will make repeated use of the two following facts. Recall that, for a fixed $i \in \mathcal R_n$, the random variable $\bX_i^{\star}=(X_{i1}^{\star}, \hdots,X_{id}^{\star})$ is defined by
\begin{equation*}
X_{ij}^{\star}=
\left \{ \begin{array}{cl}
X_{ij}& \mbox{if } j \in M\\
0 & \mbox{otherwise},
\end{array}
\right .
\end{equation*}
and $\bX_i^{(n)}= \bX_i^{\star}$ as soon as $M\subset M_{i}^{n+1-i}$. Recall also that, by definition, $\|\bX_i^{\star}\|\geq 1$.
 \begin{fact}
 \label{FAIT1}
For each $i \in \mathcal R_n$,
\begin{equation*}
\label{eq:lienSd}
S(\bX^{\star},\bX_i^{\star})=\bar S(\bX^{\star},\bX_i^{\star})=\cos(\bX^{\star},\bX_i^{\star})=1-\frac{1}{{2}}\,\de^2\left( \frac{\bX^{\star}}{\|\bX^{\star}\|},\frac{\bX_i^{\star}}{\|\bX_i^{\star}\|}\right),
\end{equation*}
where $\de$ is the usual Euclidean distance on $\R^d$.
\end{fact}
\begin{fact}
\label{FAIT2}
Let, for all $i\geq 1$, 
$$T_i=\min(k\geq i:M_i^{k+1-i}\supset M)$$
be the first time instant when user $i$ has rated all the films indexed by $M$. Set
\begin{equation}
\label{eq:defLn}
\L_n=\{i\in \mathcal R_n:T_i\leq n\},
\end{equation}
and define, for $i\in \mathcal L_n$,
$$W_{ni}^\star({\bf x}^{\star})=\left\{
\begin{array}{ll}
 1/k_n & \textrm{if } {\bf X}_i^\star\textrm{ is among the $k_n$-MS of }{\bf x}^{\star} \textrm{ in } \{{\bf X}_i^\star, i\in \L_n\} \\
0 & \textrm{otherwise}.
\end{array}\right.
$$
Then
$$W_{ni}^\star({\bf x^\star})=\left\{
  \begin{array}{ll}
1/k_n & \textrm{ if }\frac{{\bf X}_i^\star}{\|{\bf X}_i^\star\|}\textrm{ is among the $k_n$-NN of }\frac{{\bf x^{\star}}}{\|{\bx^{\star}}\|} \textrm{ in } \left\{\frac{{\bf X}_i^\star}{\|{\bf X}_i^\star\|},i\in\L_n\right\}\\
0 & \textrm{otherwise,}
  \end{array}\right.
$$
where the $k_n$-NN are evaluated with respect to the Euclidean distance on $\mathbb R^d$. That is, the $W_{ni}^\star({\bf x^\star})$ are the usual Euclidean NN weights (Gy\"orfi et al. \cite{gykokrha}), indexed by the random set $\L_n$. 
\end{fact}
Recall that $|\mathcal R_n|$ represents the number of users who have already provided information about the variable of interest (the movie Titanic in our example) at time $n$. We are now in a position to state the main result of this section.

\begin{theo}
\label{convergence}
Suppose that $\card{M}\geq 2$ and that assumption $(\mathbf F)$ is satisfied. Suppose that $k_n \to \infty$, $|\mathcal R_n|\to \infty$ a.s. and $\mathbb E[k_n/|\mathcal R_n|] \to 0$ as $n\to \infty$. Then
$$\mathbb E \left | \eta_n(\bf X^{\star})-\eta(\bf X^{\star})\right|\to 0 \quad \mbox{as } n \to \infty.$$
\end{theo}
Thus, to achieve consistency, the number of nearest neighbors $k_n$, over which one averages in order to estimate the regression function, should on the one hand tend to infinity but should, on the other hand, be small with respect to the cardinality of the subset of database users who have already rated the item of interest. We illustrate this result by working out two examples.
\begin{exmp}
\label{exemple1}
Consider, to start with, the somewhat ideal situation where all users in the database have rated the item of interest. In this case, $\mathcal R_n=\{1,\hdots,n\}$, and the asymptotic conditions on $k_n$ become $k_n\to \infty$ and $k_n/n\to 0$ as $n\to\infty$. These are just the well-known conditions ensuring consistency of the usual (i.e., Euclidean) NN regression estimate (Gy\"orfi et al. \cite{gykokrha}, Chapter 6).
\end{exmp}
\begin{exmp}
\label{exemple2}
In this more sophisticated model, we recursively define the sequence $(\mathcal {R}_n)_n$ as follows. Fix, for simplicity, $\mathcal {R}_1=\{1\}$. At step $n\geq 2$, we first decide or not to add one element to $\mathcal {R}_{n-1}$ with probability $p\in (0,1)$, independently of the data. If we decide to increase $\mathcal R_n$, then we do it by picking a random variable $B_n$ uniformly over the set $\{1,\hdots,n\}-\mathcal {R}_{n-1}$, and set $\mathcal R_n=\mathcal R_{n-1}\cup\{B_n\}$; otherwise, $\mathcal R_n=\mathcal R_{n-1}$. Clearly, $|\mathcal {R}_n|-1$ is a sum of $n-1$ independent Bernoulli random variables with parameter $p$, and it has therefore a binomial distribution with parameters $n-1$ and $p$. Consequently, 
$$\esp\left[\frac{k_n}{\card{\mathcal R_n}}\right]=\frac{k_n \left [1-(1-p)^n\right]}{np}.$$
In this setting, consistency holds provided $k_n \to \infty$ and $k_n=\mbox{o}(n)$ as $n\to \infty$.
\end{exmp}

In the sequel, the letter $C$ will denote a positive constant, the value of which may vary from line to line. Proof of Theorem \ref{convergence} will strongly rely on facts \ref{FAIT1}, \ref{FAIT2} and the following proposition. 
\begin{pro}
\label{pro:dec_eta}
Suppose that ${\card M}\geq 2$ and that assumption $(\mathbf F)$ is satisfied. Let $\alpha_{ni}=\prob(M^{n+1-i}\not\supset M\,|\,M)$. Then
\begin{align*}
& \esp \left |\eta_n({\bf X}^\star)-\eta({\bf X}^\star) \right |\\
& \quad \leq C \Bigg\{\esp\left[\frac{k_n}{\card{\mathcal R_n}}\right]  +\esp \left[\frac{1}{\card{\mathcal R_n}}\sum_{i\in \mathcal R_n}\esp \alpha_{ni} \right] +\esp\left[\prod_{i\in \mathcal R_n}\alpha_{ni}\right]\\
& \qquad + \esp\Big|\sum_{i\in\L_n}W_{ni}^\star({\bf X}^\star)\frac{Y_i}{\|{\bf X}_i^\star\|}-\varphi({\bf X}^\star)\Big|\Bigg\},
\end{align*}
where $\mathcal R_n$ stands for the non-empty subset of users who have already provided information about the variable of interest at time $n$ and $\L_n$ is defined in \eqref{eq:defLn}.
\end{pro}
{\bf Proof of Proposition \ref{pro:dec_eta}.} \quad Since $\|{\bf X}^\star\|\leq s\sqrt{d}$, it will be enough to upper bound the quantity
$$\esp\left |\sum_{i\in \mathcal R_n}W_{ni}({\bf X}^\star)\frac{Y_i}{\|{\bf X}_i^{(n)}\|}-\varphi({\bf X}^\star)\right|.$$
To this aim, we write
\begin{align*}
 & \esp\left|\sum_{i\in \mathcal R_n}W_{ni}({\bf X}^\star)\frac{Y_i}{\|{\bf X}_i^{(n)}\|}-\varphi({\bf X}^\star)\right|\\
& \quad \leq \esp\left[ \sum_{i\in \L_n^c}W_{ni}({\bf X}^\star)\frac{Y_i}{\|{\bf X}_i^{(n)}\|}\right]+\esp\left |\sum_{i\in \L_n}W_{ni}({\bf X}^\star)\frac{Y_i}{\|{\bf X}_i^{(n)}\|}-\varphi({\bf X}^\star)\right|,
\end{align*}
where the symbol $A^c$ denotes the complement of the set $A$.
Let the event
$$\mathcal A_n=\left[\exists i\in\L_n^c:{\bf X}_i^{(n)}\textrm{ is among the $k_n$-MS of }{\bf X}^\star\textrm{ in }\{{\bf X}_i^{(n)}, i\in \mathcal R_n\}\right].$$
Since $\sum_{i\in\L_n^c}W_{ni}({\bf X}^\star)\leq 1$, we have
$$\esp \left[\sum_{i\in \L_n^c}W_{ni}({\bf X}^\star)\frac{Y_i}{\|{\bf X}_i^{(n)}\|}\right] =\esp\left[\sum_{i\in \L_n^c}W_{ni}({\bf X}^\star)\frac{Y_i}{\|{\bf X}_i^{(n)}\|}\ind_{\mathcal A_n}\right]\leq s\prob(\mathcal A_n).$$
Observing that, for $i\in\L_n$, ${\bf X}_i^{(n)}={\bf X}_i^\star$ and $W_{ni}({\bf X}^\star)\ind_{\mathcal A_n^c}=W_{ni}^\star({\bf X}^\star)\ind_{\mathcal A_n^c}$ (fact \ref{FAIT2}), we obtain
\begin{align*}
& \esp  \left|\sum_{i\in \L_n}W_{ni}({\bf X}^\star)\frac{Y_i}{\|{\bf X}_i^{(n)}\|}-\varphi({\bf X}^\star)\right| \\
&  = \esp\left|\sum_{i\in \L_n}W_{ni}({\bf X}^\star)\frac{Y_i}{\|{\bf X}_i^\star\|}-\varphi({\bf X}^\star)\right| \\
&  = \esp\left|\sum_{i\in \L_n}W_{ni}({\bf X}^\star)\frac{Y_i}{\|{\bf X}_i^\star\|}-\varphi({\bf X}^\star)\right|\ind_{\mathcal A_n}+\esp\left|\sum_{i\in \L_n}W_{ni}^{\star}({\bf X}^\star)\frac{Y_i}{\|{\bf X}_i^\star\|}-\varphi({\bf X}^\star)\right|\ind_{\mathcal A_n^c} \\
&  \leq s\prob(\mathcal A_n)+\esp\left|\sum_{i\in \L_n}W_{ni}^{\star}({\bf X}^\star)\frac{Y_i}{\|{\bf X}_i^\star\|}-\varphi({\bf X}^\star)\right|.
\end{align*}
Applying finally Lemma \ref{lem:probAn} completes the proof of the proposition.
\begin{flushright}
$\square$
\end{flushright}
We are now in a position to prove Theorem \ref{convergence}.
\bigskip

{\bf Proof of Theorem \ref{convergence}.}\quad According to Proposition \ref{pro:dec_eta}, Lemma \ref{lem:Lnvide} and Lemma \ref{lem:LnsurRn}, the result will be proven if we show that $$\esp\left|\sum_{i\in \L_n}W_{ni}^\star({\bf X}^\star)\frac{Y_i}{\|{\bf X}_i^\star\|}-\varphi({\bf X}^\star)\right|\to 0\quad\textrm{as}\quad n\to\infty.$$
For $L_n\in\mathcal P(\{1,\hdots,n\})$, set 
\begin{align*}
& Z_{L_n}^n =\frac{1}{k_n}\sum_{i\in L_n}\ind_{\left[\frac{{\bf X}_i^\star}{\|{\bf X}_i^\star\|}\textrm{ is among the $k_n$-NN of }\frac{{\bf X}^\star}{\|{\bf X}^\star\|} \textrm{ in } \left\{\frac{{\bf X}_i^\star}{\|{\bf X}_i^\star\|}, i\in L_n\right\}\right]}\frac{Y_i}{\|{\bf X}_i^\star\|}-\varphi({\bf X}^\star).
\end{align*}
Conditionally on the event $[M=m]$, the random variables ${\bf X}^\star$ and $\{{\bf X}_i^\star, i\in L_n\}$ are independent and identically distributed. Thus, applying Theorem 6.1 in \cite{gykokrha}, we obtain
$$\forall\varepsilon>0,\ \exists A_m\geq 1:k_n\geq A_m\ \mbox{and} \ \frac{\card{L_n}}{k_n}\geq A_m\Longrightarrow\esp_m|Z_{L_n}^n|\leq\varepsilon,$$
where we use the notation $\esp_m[.]=\esp[.|M=m]$. Let $\prob_m(.)=\prob(.|M=m)$.  By independence,
$$\esp_m|Z_{\mathcal L_n}^n|  =  \sum_{L_n\in\mathcal P(\{1,\hdots,n\})}\esp_m|Z_{L_n}^n|\,\prob_m(\L_n=L_n).$$
Consequently, letting $A=\max A_m$, where the maximum is taken over all possible choices of $m \in \mathcal P^{\star}(\{1, \hdots, d\})$ we get, for all $n$ such that $k_n\geq A$,
\begin{align*}
\esp_m|Z_{\mathcal L_n}^n| & =  \sum_{\substack{L_n\in\mathcal P(\{1,\hdots,n\}) \\ \card{L_n}\geq Ak_n}} \esp_m|Z_{L_n}^n|\,\prob_m(\L_n=L_n)\\
& \quad +\sum_{\substack{L_n\in\mathcal P(\{1,\hdots,n\}) \\ \card{L_n}< Ak_n}} \esp_m|Z_{L_n}^n|\,\prob_m(\L_n=L_n) \\
&\leq  \varepsilon+s\prob_m(\card{\L_n}<Ak_n). 
\end{align*}
Therefore
$$\esp |Z_{\L_n}^n|=\esp \left[\esp \left[ |Z_{\L_n}^n| \,\big|\,M\right]\right] \leq \varepsilon + s \prob \left(\card{\L_n}<Ak_n\right).$$
Moreover, by Lemma \ref{lem:LnsurRn}, 
$$\frac{\card{\L_n}}{k_n}=\frac{\card{\mathcal R_n}}{k_n}\left(1- \frac{\card{\L_n^c}}
{\card{{\mathcal R}_n} }\right)\to\infty\ \textrm{in probability as } n\to \infty.$$
Thus, for all $\varepsilon >0$, $\limsup_{n\to\infty}\esp|Z_{\L_n}^n|\leq\varepsilon$, whence $\esp|Z_{\L_n}^n|\to 0$ as $n\to\infty$. This shows the desired result.
\begin{flushright}
$\square$
\end{flushright}
\section{Rates of convergence}
In this section, we bound the rate of convergence of $\mathbb E \left |\eta_n(\mathbf X^{\star})-\eta(\mathbf X^{\star}) \right|$ for the cosine-type $k_n$-NN regression estimate. To reach this objective, we will require that the function 
$$\varphi(\bx^\star)= \mathbb E\left[ \frac{Y}{\|{\bf X}^\star\|} \Big| \frac{{\bf X}^\star}{\|{\bf X}^\star\|}=\frac{{\bf x}^\star}{\|{\bf x}^\star\|} \right]$$
satisfies a Lipschitz-type property with respect to the similarity $\bar S$.  More precisely, we say that $\varphi$ is Lipschitz with respect to $\bar S$ if there exists a constant $C>0$ such that, for all $\bx$ and $\bx'$ in $\R^d$,
$$|\varphi(\bx)-\varphi(\bx')|\leq C \sqrt {1-{\bar S}(\bx,\bx')}.$$
In particular, for $\bx$ and $\bx' \in \mathbb R^d-{\bf 0}$  with the same null components, this property can be rewritten as
$$|\varphi(\bx)-\varphi(\bx')|\leq \frac{C}{\sqrt 2}\,  {\rm d} \left( \frac{\bx}{\|\bx\|},\frac{\bx'}{\|\bx'\|}\right),$$
where we recall that ${\rm d}$ denotes Euclidean distance.
\begin{theo}
\label{vitesse}
Suppose that assumption $(\mathbf F)$ is satisfied and that $\varphi$ is Lipschitz with respect to ${\bar S}$. Let $\alpha_{ni}=\prob(M^{n+1-i}\not\supset M\,|\,M)$, and assume that $\card{M}\geq 4$. Then there exists $C>0$ such that, for all $n \geq 1$,
\begin{align*}
& \mathbb E \left |\eta_n(\mathbf X^{\star})-\eta(\mathbf X^{\star}) \right|\\
& \quad \leq C \left\{\mathbb E \left[ \frac{k_n}{|\mathcal R_n|}\sum_{i \in \mathcal R_n} \mathbb \esp\,  \alpha_{ni}\right] +\esp\left[\prod_{i\in \mathcal R_n}\alpha_{ni}\right]+\mathbb E \left[\left(\frac{k_n}{|\mathcal R_n|}\right)^{P_n}\right]+\frac{1}{\sqrt{k_n}}\right\},
\end{align*}
where $P_n=1/(\card{M}-1)$ if $k_n\leq \card{\mathcal R_n}$, and $P_n=1$ otherwise.
\end{theo}
To get an intuition on the meaning of Theorem \ref{vitesse}, it helps to note that the terms depending on $\alpha_{ni}$ do measure the influence of the unrated items on the performance of the estimate. Clearly, this performance improves as the $\alpha_{ni}$ decrease, i.e., as the proportion of rated items growths. On the other hand, the term $\mathbb E  [ ({k_n}/{|\mathcal R_n|} )^{P_n}]$ can be interpreted as a bias term in dimension $|M|-1$, whereas ${1}/{\sqrt{k_n}}$ represents a variance term. As usual in nonparametric estimation, the rate of convergence of the estimate is dramatically deteriorated as $|M|$ becomes large. However, in practice, this drawback may be circumvented by using preliminary dimension reduction steps, such as factorial methods (PCA, etc.) or inverse regression methods (SIR, etc.).

\begin{exmp}({\bf cont. Example \ref{exemple1}}) Recall that we assume, in this ideal model, that $\mathcal R_n=\{1, \hdots, n\}$. Suppose in addition that $M=\{1,\hdots,d\}$, i.e., any new user in the database rates all products the first time he enters the database. Then the upper bound of Theorem \ref{vitesse} becomes
$$ \mathbb E \left |\eta_n(\mathbf X^{\star})-\eta(\mathbf X^{\star}) \right |=  \emph{O} \left( \left(\frac{k_n}{n}\right)^{1/(d-1)}+\frac{1}{\sqrt{k_n}}\right).$$
Since neither $\mathcal R_n$ nor $M$ are random in this model, we see that there is no influence of the dynamical rating process. Besides, we recognize the usual rate of convergence of the Euclidean NN regression estimate (Gy\"orfi et al. \cite{gykokrha}, Chapter 6) in dimension $d-1$. In particular, the choice $k_n\sim n^{2/(d+1)}$ leads to 
$$\mathbb E \left |\eta_n(\mathbf X^{\star})-\eta(\mathbf X^{\star}) \right|=\emph{O}\left (n^{-1/(d+1)}\right).$$
Note that we are led to a $d-1$-dimensional rate of convergence (instead of the usual $d$) just because everything happens as if the data is projected on the unit sphere of $\mathbb{R}^d$. \end{exmp}

\begin{exmp}({\bf cont. Example \ref{exemple2}}) In addition to model \ref{exemple2}, we suppose that at each time, a user entering the game reveals his preferences according to the following sequential procedure. At time 1, the user rates exactly 4 items by randomly guessing in $\{1,\hdots,d\}$. At time $2$, he updates his preferences by adding exactly one rating among his unrated items, randomly chosen in $\{1,\hdots,d\}-M_1^1$. Similarly, at time $3$, the user revises his preferences according to a new item uniformly selected in $\{1,\hdots,d\}-M_1^2$, and so on. In such a scenario, $|M^j|=\min(d,j+3)$ and thus, $M^j=\{1,\hdots,d\}$ for $j\geq d-3$. Moreover, since $|M|=4$, a moment's thought shows that
$$\alpha_{ni}=\left\{
  \begin{array}{ll}
0 & \textrm{if }i\leq n-d+4 \\
1-\frac{\displaystyle \binom{d-4}{n-i}}{\displaystyle \binom{d}{n+4-i}} & \textrm{if } n-d+5\leq i\leq n.
  \end{array}\right.
$$
Assuming $n\geq d-5$, we obtain 
\begin{align*}
 \sum_{i\in\mathcal R_n}\alpha_{ni} &  \leq \sum_{i=n-d+5}^n\alpha_{ni}\\
&\leq\sum_{i=n-d+5}^n\left(1-\frac{(n+4-i)(n+3-i)(n+2-i)(n+1-i)}{d(d-1)(d-2)(d-3)}\right) \\
& \leq (d-4)\left(1-\frac{24}{d(d-1)(d-2)(d-3)}\right).
\end{align*}
Similarly, letting $\mathcal {R}_{n0}=\mathcal R_n\cap \{n-d+5,\hdots,n\}$, we have
\begin{align*}
\prod_{i\in\mathcal R_n}\alpha_{ni}& = \prod_{i\in\mathcal {R}_{n0}}\alpha_{ni}\, \ind_{\{\min(\mathcal R_n)\geq n-d+5\}}\\
& \leq\left(1-\frac{24}{d(d-1)(d-2)(d-3)}\right)^{\card{\mathcal {R}_{n0}}}\ind_{\{\min(\mathcal R_n)\geq n-d+5\}}.
\end{align*}
Since $|{\mathcal R}_n|-1$ has binomial distribution with parameters $n-1$ and $p$, we obtain
\begin{eqnarray*}
\esp\left[\prod_{i\in \mathcal R_n}\alpha_{ni}\right] &\leq & \mathbb{P}\big(\min(\mathcal R_n)\geq n-d+5\big)\\
& \leq & \mathbb{P}\big(|\mathcal {R}_n|\leq d-5\big) \leq  \frac{C}{n}.
\end{eqnarray*}
Finally, applying Jensen's inequality,
\begin{eqnarray*}
\mathbb E \left[\left(\frac{k_n}{|\mathcal R_n|}\right)^{P_n}\right] & = & \mathbb E \left[ \left (\frac{k_n}{|\mathcal R_n|}\right)^{1/3}{\bf 1}_{\{k_n\leq |\mathcal {R}_n|\}}\right]+\mathbb E \left[ \frac{k_n}{|\mathcal R_n|}{\bf 1}_{\{k_n > |\mathcal {R}_n|\}}\right]\\
 & \leq & C \left (\mathbb E \left[ \frac{k_n}{|\mathcal R_n|}\right] \right)^{1/3}\leq C \left(\frac{k_n}{n}\right)^{1/3}.
 \end{eqnarray*}
Putting all the pieces together, we get with Theorem \ref{vitesse}
$$\mathbb E \left |\eta_n(\mathbf X^{\star})-\eta(\mathbf X^{\star}) \right| = \emph{O} \left ( \left(\frac{k_n}{n}\right)^{1/3}+\frac{1}{\sqrt{k_n}}\right).$$
In particular, the choice $k_n \sim n^{2/5}$ leads to 
$$\mathbb E \left |\eta_n(\mathbf X^{\star})-\eta(\mathbf X^{\star}) \right| =\emph{O}(n ^{-1/5}),$$
which is the usual NN regression estimate rate of convergence when the data is projected on the unit sphere of $\mathbb {R}^4$.
\end{exmp}
{\bf Proof of Theorem \ref{vitesse}.}\quad Starting from Proposition \ref{pro:dec_eta}, we just need to upper bound the quantity
$$\esp\left|\sum_{i\in\L_n}W_{ni}^\star({\bf X}^\star)\frac{Y_i}{\|{\bf X}_i^\star\|}-\varphi({\bf X}^\star)\right|.$$
A combination of  Lemma \ref{lem:rate1ppv} and the proof of Theorem 6.2 in \cite{gykokrha} shows that
\begin{align}
& \esp\bigg|\sum_{i\in\L_n}W_{ni}^\star({\bf X}^\star)\frac{Y_i}{\|{\bf X}_i^\star\|}-\varphi({\bf X}^\star)\bigg|\nonumber\\
& \quad \leq C\left\{\frac{1}{\sqrt{k_n}}+\esp \left[\left(\frac{k_n}{\card{\L_n}}\right)^{1/(\card{M}-1)}{\bf 1}_{\{\L_n\neq \emptyset\}}\right]+\prob(\L_n=\emptyset)\right\}.\label{ineq:M}
\end{align}
We obtain 
\begin{align*}
& \esp\left[\left(\frac{k_n}{\card{\L_n}}\right)^{1/(\card{M}-1)}{\bf 1}_{\{\L_n\neq \emptyset\}}\right]\\
& \quad =\esp \left[ \left(\frac{k_n}{{\card{{\cal R}_n}}\big(1-\card{\L_n^c}/\card{{\cal R}_n}\big)}\right)^{1/({\card M}-1)}{\bf 1}_{\{\card{\L_n^c}\leq \card{{\cal R}_n}/2\}}\right]\\
&  \qquad + \esp \left[ \left(\frac{k_n}{{\card{\L_n}}}\right)^{1/({\card M}-1)}{\bf 1}_{\{\card{\L_n^c}> \card{{\cal R}_n}/2\}} {\bf 1}_{\{\L_n\neq \emptyset\}}\right]\\
& \quad \leq  \esp \left[ \left( \frac{2k_n}{{\card{\mathcal R_n}}}\right)^{1/(\card{M}-1)}\right]+\esp\left[ k_n^{1/(\card{M}-1)} {\bf 1}_{\{\card{\L_n^c}>\card {\mathcal R_n}/2\}}\right].
\end{align*}
Since $\card{M}\geq 4$, one has $2^{1/({\card M}-1)}\leq 2$ and $k_n^{1/(\card{M}-1)} \leq k_n$ in the rightmost term, so that, thanks to Lemma 
\ref{lem:LnsurRn},
\begin{align*}
& \esp\left[\left(\frac{k_n}{\card{\L_n}}\right)^{1/(\card{M}-1)}{\bf 1}_{\{\L_n\neq \emptyset\}}\right]\\
& \quad \leq C
\left\{ \esp \left[ \left( \frac{k_n}{{\card{\mathcal R_n}}}\right)^{1/(\card{M}-1)}\right]+\esp\left[\frac{k_n}{\card{\mathcal R_n}}\sum_{i\in \mathcal R_n}\esp\,  \alpha_{ni}\right]\right\}.
\end{align*}
The theorem is a straightforward combination of Proposition \ref{pro:dec_eta}, inequality (\ref{ineq:M}), and Lemma \ref{lem:Lnvide}.
\begin{flushright}
$\square$
\end{flushright}
\section{Technical lemmas}

Before stating some technical lemmas, we remind the reader that $\mathcal R_n$ stands for the non-empty subset of $\{1,\hdots,n\}$ of users who have already rated the variable of interest at time $n$. Recall also that, for all $i\geq 1$,
$$T_i=\min(k\geq i:M_i^{k+1-i}\supset M)$$
and
$$\L_n=\{i\in \mathcal R_n:T_i\leq n\}.$$

\begin{lem}
\label{lem:Lnvide}
We have
$$\prob(\L_n=\emptyset)=\esp\left[\prod_{i\in \mathcal R_n}\alpha_{ni}\right]\to 0 \quad \mbox{as} \ n\to\infty.$$
\end{lem}
{\bf Proof of Lemma \ref{lem:Lnvide}.}\quad Conditionally on $M$ and ${\cal R}_n$, the random variables $\{T_i,i\in {\cal R}_n\}$ are independent. Moreover, the sequence $(M^n)_{n\geq 1}$ is nondecreasing. Thus, the identity $[T_i>n]=[M_i^{n+1-i}\not\supset M]$ holds for all $i\in\mathcal R_n$. Hence, 
\begin{align*}
\prob(\L_n=\emptyset) & =\prob\left(\forall i\in \mathcal R_n:T_i>n\right)\\
& =\esp\left[\prob\left(\forall i\in \mathcal R_n:T_i>n\,\Big |\,\mathcal R_n,M\right)\right] \\
 & =\esp\left[\prod_{i\in \mathcal R_n}\prob\left(T_i>n\,\Big |\,\mathcal R_n,M\right)\right]\\
 &=\esp\left[\prod_{i\in \mathcal R_n}\prob\left(M_i^{n+1-i}\not\supset M\,\Big |\,M\right)\right] \\
 & \quad \mbox{(by independence of $(M_i^{n+1-i},M)$  and $\mathcal R_n$}\\ 
 & =\esp\left[\prod_{i\in \mathcal R_n}\alpha_{ni}\right].
\end{align*}
The last statement of the lemma is clear since, for all $i$, $\alpha_{ni}\to 0$ a.s. as $n\to\infty$. 
\begin{flushright}$\square$\end{flushright}

\begin{lem}
\label{lem:LnsurRn}  We have
$$
\esp\left[\frac{\card{\L_n^c}}{\card{\mathcal R_n}}\right]=\esp\left[\frac{1}{\card{\mathcal R_n}}\sum_{i\in \mathcal R_n}\esp\,  \alpha_{ni}\right]$$
and  
$$\esp\left[ \frac{1}{\card{\L_n}}{\bf 1}_{\{\L_n\neq \emptyset\}}\right]\leq 2\esp\left[\frac{1}{\card{{\cal R}_n}}\right]+2\esp\left[\frac{1}{\card{\mathcal R_n}}\sum_{i\in \mathcal R_n}\esp\, \alpha_{ni}\right].$$

Moreover, if $\lim_{n\to\infty}\card{\mathcal R_n}=\infty$ a.s., then
$$\lim_{n\to\infty}\esp\left[\frac{\card{\L_n^c}}{\card{\mathcal R_n}}\right]= 0.$$
\end{lem}
{\bf Proof of Lemma \ref{lem:LnsurRn}.}\quad First, using the fact that the sequence $(M^n)_{n\geq 1}$ is nondecreasing, we see that for all $i\in \mathcal R_n$, $[T_i>n]=[M_i^{n+1-i}\not\supset M]$. Next, recalling that $\mathcal R_n$ is independent of $T_i$ for fixed $i$, we obtain
$$\esp\left[\frac{\card{\L_n^c}}{\card{\mathcal R_n}}\,\Big |\,\mathcal R_n\right]=\frac{1}{\card{\mathcal R_n}}\esp\left[\sum_{i\in \mathcal R_n}\ind_{\{T_i>n\}}\,\Big|\,\mathcal R_n\right]=\frac{1}{\card{\mathcal R_n}}\sum_{i\in \mathcal R_n}\prob(M_i^{n+1-i}\not\supset M),$$
and this proves the first statement of the lemma. Now define $\mathcal J_n=\{n+1-i,i\in \mathcal R_n\}$ and observe that
$$\esp \left [ \frac{\card{\L_n^c}}{\card{\mathcal R_n}}\right]=\esp\left[\frac{1}{\card{\mathcal J_n}}\sum_{j\in \mathcal J_n}\prob(M^j\not\supset M)\right],$$
where we used $|\mathcal J_n|=|\mathcal R_n|$. Since, by assumption, $\card{\mathcal J_n}=\card{\mathcal R_n}\to\infty$ a.s. as $n \to\infty$ and $\prob(M^j\not\supset M)\to 0$ as $j\to\infty$, we obtain
$$\lim_{n\to\infty}\frac{1}{\card{\mathcal J_n}}\sum_{j\in \mathcal J_n}\prob(M^j\not\supset M)=0\quad \mbox{a.s.}$$
The conclusion follows by applying Lebesgue's dominated convergence Theorem. The second statement of the lemma is obtained from the following chain of inequalities: 
\begin{eqnarray*}
\esp\left[ \frac{1}{\card{\L_n}}{\bf 1}_{\{\L_n\neq \emptyset\}}\right]
& = & \esp \left[ \frac{1}{{\card{{\cal R}_n}}\big(1-\card{\L_n^c}/\card{{\cal R}_n}\big)}{\bf 1}_{\{\L_n\neq \emptyset\}}\right]\\
& = & \esp \left[ \frac{1}{{\card{{\cal R}_n}}\big(1-\card{\L_n^c}/\card{{\cal R}_n}\big)}{\bf 1}_{\{\card{\L_n^c}\leq \card{{\cal R}_n}/2\}}\right]\\
& & + \esp \left[ \frac{1}{{\card{\L_n}}}{\bf 1}_{\{\card{\L_n^c}> \card{{\cal R}_n}/2\}} {\bf 1}_{\{\L_n\neq \emptyset\}}\right]\label{eq:Ln}\\
& \leq & 2\esp \left[ \frac{1}{\card{{\cal R}_n}}\right]+\prob\left(\card{\L_n^c}>\frac{\card{{\cal R}_n}}{2}\right)\\
& \leq & 2\esp \left[ \frac{1}{\card{{\cal R}_n}}\right]+2\esp\left[\frac{\card{\L_n^c}}{\card{\mathcal R_n}}\right].
\end{eqnarray*}
Applying the first part of the lemma completes the proof. 
\begin{flushright}
$\square$
\end{flushright}

\begin{lem}
\label{lem:beurk} Denote by $\mathbf Z^\star$ and $\mathbf Z_1^\star$ the random variables  $\mathbf Z^\star=\bX^\star/\|\bX^\star\|$, $\mathbf Z_1^\star=\bX_1^\star/\|\bX_1^\star\|$, and let $\xi(\mathbf Z^{\star})=\prob(S(\mathbf Z^{\star},\mathbf Z_1^\star)>1/2\,|\,\mathbf Z^{\star})$. Then 
\begin{align*}
\prob\Big(2k_n>\card{\L_n}\xi(\mathbf Z^\star)\,\big|\,\L_n,M\Big)\leq & \,2\esp \left[\frac{k_n}{\card{{\cal R}_n}}\,\Big|\,\L_n\right] \esp\left[\frac{1}{\xi(\mathbf Z^\star)}\,\Big|\,M\right] \\
& +\esp\left[\frac{\card{\L_n^c}}{\card{{\cal R}_n}}\,\Big|\,\L_n,M\right].
\end{align*}
\end{lem}
{\bf Proof of Lemma \ref{lem:beurk}.}\quad If $M$ is fixed, $\mathbf Z^\star$ is independent of $\L_n$ and $\mathcal R_n$. Thus, by Markov's inequality, 
\begin{align*} 
&  \prob\Big(2k_n>\card{\L_n}\xi(\mathbf Z^\star)\,\big|\,\L_n,M,{\cal R}_n\Big) \\
& \quad  = \prob\Big(2k_n> \card{{\cal R}_n}\xi(\mathbf Z^\star)-\card{\L_n^c}\xi(\mathbf Z^\star)\,\big|\,\L_n,M,{\cal R}_n\Big)\\
 &  \quad = \prob\Big(2k_n+ \card{\L_n^c}\xi(\mathbf Z^\star)\geq \card{{\cal R}_n} \xi(\mathbf Z^\star)\,\big|\, \L_n,M,{\cal R}_n\Big)\\
 & \quad  \leq \frac{2k_n}{\card{{\cal R}_n}} \esp\left[ \frac{1}{\xi(\mathbf Z^\star)}\,\Big|\, M\right]+\frac{\card{\L_n^c}}{\card{{\cal R}_n}}.
 \end{align*}
The proof is completed by observing that ${\cal R}_n$ and $M$ are independent random variables. 
\begin{flushright}
$\square$
\end{flushright}

Let $\mathcal B({\bf x},\varepsilon)$ be the closed Euclidean ball in $\R^d$ centered at ${\bf x}$ of radius $\varepsilon$. Recall that the support of a probability measure $\mu$ is defined as the closure of the collection of all $\bx$ with $\mu(\mathcal B({\bf x},\varepsilon))>0$ for all $\varepsilon>0$. The next lemma can be proved with a slight modification of the proof  of Lemma 10.2 in Devroye et al. \cite{DGL}.
\begin{lem}
\label{lem:intunsurmu}
Let $\mu$ be a probability measure on $\R^d$ with a compact support. Then
$$\int \frac{1}{\mu(\mathcal B({\bf x},r))}\mu(\mathrm{d}{\bf x})\leq C,$$
with $C>0$ a constant depending upon $d$ and $r$ only.
\end{lem}

\begin{lem}
\label{lem:probAn}
Suppose that $\card{M}\geq 2$, and let the event
$$\mathcal A_n=\left [\exists i\in\L_n^c:{\bf X}_i^{(n)}\textrm{ is among the $k_n$-MS of }{\bf X}^\star\textrm{ in }{\{\bf X}_i^{(n)},i\in \mathcal R_n\}\right].$$
Then
$$\prob(\mathcal A_n)\leq C\left\{\mathbb E \left[\frac{k_n}{\card{\mathcal R_n}}\right]+\esp\left[\frac{1}{\card{\mathcal R_n}}\sum_{i\in \mathcal R_n}\esp \alpha_{ni}\right]+\esp\left[\prod_{i\in \mathcal R_n} \alpha_{ni}\right]\right\}.$$
\end{lem}
{\bf Proof of Lemma \ref{lem:probAn}.}\quad Recall that, for a fixed $i \in \mathcal R_n$, the random variable $\bX_i^{\star}=(X_{i1}^{\star}, \hdots,X_{id}^{\star})$ is defined by
\begin{equation*}
X_{ij}^{\star}=
\left \{ \begin{array}{cl}
X_{ij}& \mbox{if } j \in M\\
0 & \mbox{otherwise},
\end{array}
\right .
\end{equation*}
and $\bX_i^{(n)}= \bX_i^{\star}$ as soon as $M\subset M_{i}^{n+1-i}$.
\bigskip

We first prove the inclusion
\begin{equation}
\label{eq:part1_lem4}
\mathcal A_n\subset\left[\card{\{j\in\L_n:S(\bX^\star,\bX_j^\star)>1/2\}}\leq k_n\right].
\end{equation}
Take $i\in\L_n^c$ such that $\bX_i^{(n)}$ is among the $k_n$-MS of $\bX^\star$ in $\{\bX_i^{(n)},i\in \mathcal R_n\}$. Then, for all $j\in\L_n$ such that $S(\bX^\star,\bX_j^\star)>1/2$, we have 
$$S(\bX^\star,\bX_j^\star)>\frac{1}{2}\geq p_i^{(n)}\bar S(\bX^\star,\bX_i^{(n)})=S(\bX^\star,\bX_i^{(n)})$$
since $p_i^{(n)}\leq 1-1/\card{M}\leq 1/2$ if $\card{M}\geq 2$. If
$$\card{\{j\in\L_n:S(\bX^\star,\bX_j^\star)>1/2\}}> k_n,$$
then $\bX_i^{(n)}$ is not among the $k_n$-MS of $\bX^\star$ among the $\{\bX_i^{(n)},i\in \mathcal R_n\}$. This contradicts the assumption on $\bX_i^{(n)}$ and proves inclusion \eqref{eq:part1_lem4}.
\bigskip

Next, define $\mathbf Z^\star=\bX^\star/\|\bX^\star\|$, $\mathbf Z_i^\star=\bX_i^\star/\|\bX_i^\star\|$, $i=1,\hdots,n$, and let $\xi(\mathbf Z^{\star})=\prob(S(\mathbf Z^{\star},\mathbf Z_1^\star)>1/2\,|\,\mathbf Z^\star)$. If $k_n-\card{\L_n}\xi(\bZ^\star)\leq -(1/2)\card{\L_n}\xi(\bZ^\star)$ and $\L_n\neq\emptyset$, we deduce from \eqref{eq:part1_lem4} that
\begin{align*}
& \mathbb P\left (\mathcal A_n \,\Big|\,\L_n,\bZ^\star\right)\\
&\quad \leq \prob   \left(\sum_{j\in\L_n}\ind_{\{S(\bZ^{\star},\bZ_j^{\star})>1/2\}}\leq k_n\,\Big|\,\L_n,\bZ^\star\right) \\
&\quad =  \prob\left(\sum_{j\in\L_n}\left(\ind_{\{S(\bZ^\star,\bZ_j^\star)>1/2\}}-\xi(\bZ^\star)\right)\leq k_n-\card{\L_n}\xi(\bZ^\star)\,\Big|\,\L_n,\bZ^\star\right) \\
& \quad \leq  \prob\left(\sum_{j\in\L_n}\left(\ind_{\{S(\bZ^\star,\bZ_j^\star)>1/2\}}-\xi(\bZ^\star)\right)\leq -\frac{1}{2}\card{\L_n}\xi(\bZ^\star)\,\Big|\,\L_n,\bZ^\star\right) \\
& \quad \leq  \frac{4\card{\L_n}\xi(\bZ^\star)}{\left(\card{\L_n}\xi(\bZ^\star)\right)^2}=\frac{4}{\card{\L_n}\xi(\bZ^\star)}\\
& \qquad \mbox {(by Tchebychev's inequality).}
\end{align*}
In the last inequality, we use the fact that, since $\sigma(M)\subset \sigma (\mathbf Z^\star)$, the random variables $\{\mathbf Z^\star_i,i\in\L_n\}$ are independent conditionally on $\mathbf Z^\star$ and $\L_n$. Using again the inclusion $\sigma(M)\subset \sigma (\mathbf Z^\star)$, we obtain, on the event $[\mathcal L_n \neq 0]$,
\begin{align*}
& \mathbb P\left (\mathcal A_n \,\Big|\,\L_n,M \right)\\
& \quad = \esp \left[ \prob \left( \mathcal A_n\Big| \L_n,\mathbf Z^\star\right)\Big|\L_n,M\right]\\
& \quad \leq  \frac{4}{\card{\L_n}}\esp\left[\frac{1}{\xi(\bZ^\star)}\,\Big|\,\L_n,M\right]+\prob\left(k_n-\card{\L_n}\xi(\bZ^\star)>-\frac{1}{2}\card{\L_n}\xi(\bZ^\star)\,\Big|\,\L_n,M\right) \\
& \quad =  \frac{4}{\card{\L_n}}\esp\left[\frac{1}{\xi(\bZ^\star)}\,\Big|\,M\right]+\prob\left(\card{\L_n}\xi(\bZ^\star)<2k_n\,\Big|\,\L_n,M\right).
\end{align*}
Applying Lemma \ref{lem:beurk}, on the event $[\L_n\neq \emptyset]$,
\begin{align*}
& \mathbb P\left (\mathcal A_n \,\Big|\,\L_n,M \right)\\
& \leq  \frac{4}{\card{\L_n}}\esp\left[\frac{1}{\xi(\bZ^\star)}\,\Big|\,M\right]+
2\esp \left[\frac{k_n}{\card{{\cal R}_n}}\,\Big|\,\L_n\right] \esp\left[\frac{1}{\xi(\mathbf Z^\star)}\,\Big|\,M\right]+\esp\left[\frac{\card{\L_n^c}}{\card{{\cal R}_n}}\,\Big|\,\L_n,M\right].
\end{align*}
Moreover, by fact \ref{FAIT1}, 
$$\xi(\bZ^\star)=\prob\left(S(\bZ^\star,\bZ_1^\star)>\frac{1}{2}\,\Big|\,\bZ^\star\right)\geq \prob\left(d^2(\bZ^\star,\bZ_1^\star)\leq \frac{1}{2}\,\Big|\,\bZ^\star\right).$$
Thus, denoting by $\nu^M$ the distribution of $\bZ^\star$ conditionally to $M$, we deduce from Lemma \ref{lem:intunsurmu} that
$$\esp\left[\frac{1}{\xi(\bZ^\star)}\,\Big|\,M\right]\leq \int \frac{1}{\nu^M(\mathcal B(\bz,1/\sqrt 2))}\nu^M(\mathrm{d} \bz)\leq C,$$
where the constant $C$ does not depend on $M$. Putting all the pieces together, we obtain
$$\prob(\mathcal A_n)\leq C\left\{ \esp\left[\frac{1}{\card{\L_n}}\ind_{\{\L_n\neq\emptyset\}}\right]+\esp \left[\frac{k_n}{\card{{\cal R}_n}}\right]+\esp \left[ \frac{\card{\L_n^c}}{\card{{\cal R}_n}}\right]\right\}+\prob(\L_n=\emptyset).$$
We conclude the proof with Lemma \ref{lem:Lnvide} and Lemma \ref{lem:LnsurRn}.
\begin{flushright}
$\square$
\end{flushright}

In the sequel, we let ${\bf X}_{(1)}^\star, \hdots, {\bf X}_{({\card {\L_n}})}^\star$ be the sequence $\{\bX_i^\star,i\in\L_n\}$ reordered according to decreasing similarities $S(\bX^{\star},{\bf X}_{i}^\star),i\in\L_n$, that is, 
$$S\left(\bX^{\star},{\bf X}_{(1)}^\star\right)\geq \hdots \geq S\left(\bX^{\star},{\bf X}_{(\card{\L_n})}^\star\right).$$ 
Lemma \ref{lem:rate1ppv} below states the rate of convergence to 1 of $S(\bX^{\star},{\bf X}_{(1)}^\star)$.
\begin{lem}
\label{lem:rate1ppv}
Suppose that $\card{M}\geq 4$. Then there exists $C>0$ such that, on the event $[\L_n\neq \emptyset]$, 
$$1-\esp\left[S({\bf X}^\star,{\bf X}_{(1)}^\star)\,|\,M,\L_n\right]\leq\frac{C}{\card{\L_n}^{2/(\card{M}-1)}}.$$
\end{lem}
{\bf Proof of Lemma \ref{lem:rate1ppv}.}\quad Observe that 
\begin{align*}
\esp & \left[ 1-S(\bX^\star,\bX_{(1)}^\star)\,|\,\bX^\star,\L_n\right] \\
& \qquad \qquad= \int_0^1 \prob \left(1-S(\bX^\star,\bX_{(1)}^\star)>\varepsilon \,\big|\,\bX^\star,\L_n\right) \mathrm{d}\varepsilon \\
& \qquad \qquad=\int_0^1 \prob\left( \forall i\in \L_n  : 1-S(\bX^\star,\bX_i^\star)>\varepsilon\,\big|\, \bX^\star,\L_n\right) \mathrm{d}\varepsilon.
 \end{align*}

 Since $\sigma(M)\subset \sigma(\bX^\star)$, given $\bX^\star$ and $\L_n$, the random variables $\{\bX_i^\star,i\in\L_n\}$ are independent and identically distributed. Hence, 
 $$\esp \left[ 1-S(\bX^\star,\bX_{(1)}^\star)\,|\,\bX^\star, \L_n\right]= \int_0^1 \left [\prob\left( 1-S(\bX^\star,\bX_1^\star)>\varepsilon\,\big|\,\bX^\star \right)\right]^{\card{\L_n}}\mathrm{d}\varepsilon.$$
Denote by $\nu^M$ the conditional distribution of $\bX^\star/\|\bX^\star\|$ given $M$. The support of $\nu^M$ is contained in both the unit sphere of $\R^d$ and in a $\card{M}$-dimensional vector space. Thus, for simplicity, we shall consider that the support of $\nu^M$ is contained in the unit sphere of $\R^{\card M}$. Let ${\cal B}^{\card M}(\bx,r)$ be the closed Euclidean ball in $\R^{\card M}$ centered at $\bx$ of radius $r$. Since $\bX^\star$ (resp. $\bX^\star_1$) only depends on $M$ and $\bX$ (resp. $\bX_1$), then, given $\bX^\star$, the random variable $\bX^\star_1/\|\bX_1^\star\|$ is distributed according to $\nu^M$. Thus, for any $\varepsilon >0$, we may write (fact \ref{FAIT1})
$$\prob\left( 1-S(\bX^\star,\bX_1^\star)>\varepsilon\,\big|\, \bX^\star \right)=1-\nu^M\left(\mathcal B^{\card M}\left(\frac{\bX^{\star}}{\|\bX^{\star}\|},\sqrt {2\varepsilon}\right)\right),$$
and, consequently,
$$\esp \left[ 1-S(\bX^\star,\bX_{(1)}^\star)\,|\,\bX^\star,\L_n\right]= \int_0^1 \left[ 1-\nu^M\left(\mathcal B^{\card M}\left(\frac{\bX^{\star}}{\|\bX^{\star}\|},\sqrt {2\varepsilon}\right)\right) \right]^{\card{\L_n}}\mathrm{d}\varepsilon.$$
 Using the inclusion $\sigma(M)\subset \sigma (\bX^\star)$, we obtain
  \begin{align}
  \label{eq:bla}
  & \esp \left[ 1-S(\bX^\star,\bX_{(1)}^\star)\,|\,M,\L_n\right]\nonumber\\
  & \quad = \int_0^1\esp  \left[ \left\{1-\nu^M\left(\mathcal B^{\card M}\left(\frac{\bX^{\star}}{\|\bX^{\star}\|},\sqrt {2\varepsilon}\right)\right) \right\}^{\card{\L_n}}\,\Big|\, M,\L_n\right]\mathrm{d}\varepsilon.
  \end{align}

Fix $\varepsilon >0$, and denote by ${\cal S}(M)$ the support of $\nu^M$. There exists Euclidean balls $A_1,\hdots,A_{N(\varepsilon)}$ in $\R^{\card{M}}$ with radius $\sqrt {2\varepsilon}/2$ such that
 $${\cal S}(M)\subset \bigcup_{j=1}^{N(\varepsilon)} A_j \quad \mbox{and} \quad N(\varepsilon)\leq \frac{C}{\varepsilon^{({\card M}-1)/2}},$$
for some $C>0$ which may be chosen independently of $M$. Clearly, if ${\bf x}\in A_j\cap {\cal S}(M)$, then $A_j\subset \mathcal B^{\card M}({\bf x},\sqrt {2\varepsilon})$. Thus,
\begin{align*}
& \esp\left[\left\{1-\nu^M\left(\mathcal B^{\card M}\left(\frac{\bX^{\star}}{\|\bX^{\star}\|},\sqrt {2\varepsilon}\right)\right)\right\}^{\card{\L_n}} \Big|M,\L_n\right] \\
& \quad \leq \sum_{j=1}^{N(\varepsilon)}\int_{A_j}\esp\left[\left\{1-\nu^M\left(\mathcal B^M\left(\frac{\bX^{\star}}{\|\bX^{\star}\|},\sqrt {2\varepsilon}\right)\right)\right\}^{\card{\L_n}} \Big|M,\L_n\right]
\nu^M(\mathrm{d}\bx) \nonumber\\
& \quad \leq \sum_{j=1}^{N(\varepsilon)}\int_{A_j}\left(1-\nu^M(A_j)\right)^{\card{\L_n}}\nu^M(\mathrm{d}\bx) \nonumber\\
& \quad \leq  \sum_{j=1}^{N(\varepsilon)}\nu^M(A_j)\left(1-\nu^M(A_j)\right)^{\card{\L_n}} \nonumber\\
& \quad \leq N(\varepsilon)\max_{t\in [0,1]}t(1-t)^{\card{\L_n}} \nonumber\\
& \quad \leq \frac{C}{\card{\L_n}\, \varepsilon^{(\card{M}-1)/2}}.
\end{align*}
Combining this inequality and equality \eqref{eq:bla}, we obtain
\begin{align*}
& \esp\left[1-S({\bf X}^\star,{\bf X}_{(1)}^\star)\,|\,M, \L_n\right]
  \leq \int_0^1\min\left(1,\frac{C}{\card{\L_n}\, \varepsilon^{(\card{M}-1)/2}} \right)\mathrm{d}\varepsilon.
\end{align*}
Since  $\card{M}\geq 4$, an easy calculation shows that there exists $C>0$ such that
$$\esp\left[1-S({\bf X}^{\star},{\bf X}_{(1)}^\star)\,|\,M,\L_n\right]\leq\frac{C}{\card{\L_n}^{2/(\card{M}-1)}},$$
which leads to the desired result.
\begin{flushright}
$\square$
\end{flushright}

\paragraph{Acknowledgments.} The authors are greatly indebted to Albert Ben\-ve\-nis\-te for pointing out this problem. They also thank Kevin Bleakley and Toby Hocking for their careful reading of the paper, and two referees and the Associate Editor for valuable comments and insightful suggestions.

\bibliographystyle{plain}
\bibliography{biblio}

\end{document}